# Deep differentiable forest with sparse attention for the tabular data

Yingshi Chen

Institute of Electromagnetics and Acoustics, and Department of Electronic Science, Xiamen University, Xiamen 361005, China
E-mail: gsp@grusoft.com

**Abstract**

    We present a general architecture of deep differentiable forest and its sparse attention mechanism. Differentiable forest has the advantages of both trees and neural networks. Its' simple structure is easy to use and explain. It has full differentiability and all variables are learnable parameters. We would train it by the gradient-based optimization method, which shows great power in training of deep CNN. We find and analyze the attention mechanism in the differentiable forest. That is, each decision depends on only a few important features, and others are irrelevant. The attention is always sparse. Based on this observation, we improve its sparsity by data-aware initialization. We use the attribute importance to initialize the attention weight. Then the learned weight is much sparse than that from random initialization. Our experiment on some large tabular dataset shows differentiable forest has higher accuracy than GBDT, which is the state of art algorithm for tabular datasets. The source codes are available at https://github.com/closest-git/QuantumForest.

## 1. Introduction

    Differentiable decision tree[1,2] is a novel structure with the advantages of both trees and neural networks. Deep neural networks are the best machine learning method now. However tree model still has some advantages. Its structure is very simple, easy to use and explain the decision process. Especially for tabular data, tree-based GBDT models usually have better accuracy than deep networks, which is verified by many kaggle competitions and real applications. While keeping simple tree structure, the differentiable trees also have full differentiability like neural networks. So we could train it with many powerful optimization algorithms (SGD, Adam,…), just like the training of deep CNN. It could use batch training to reduce memory usage greatly. It could use the end-to-end learning mode. No need of many works to preprocess data. We have implemented this method in the open source library QuantumForest. Experiments on large datasets show that QuantumForest has higher accuracy than both deep networks and best GBDT libs.

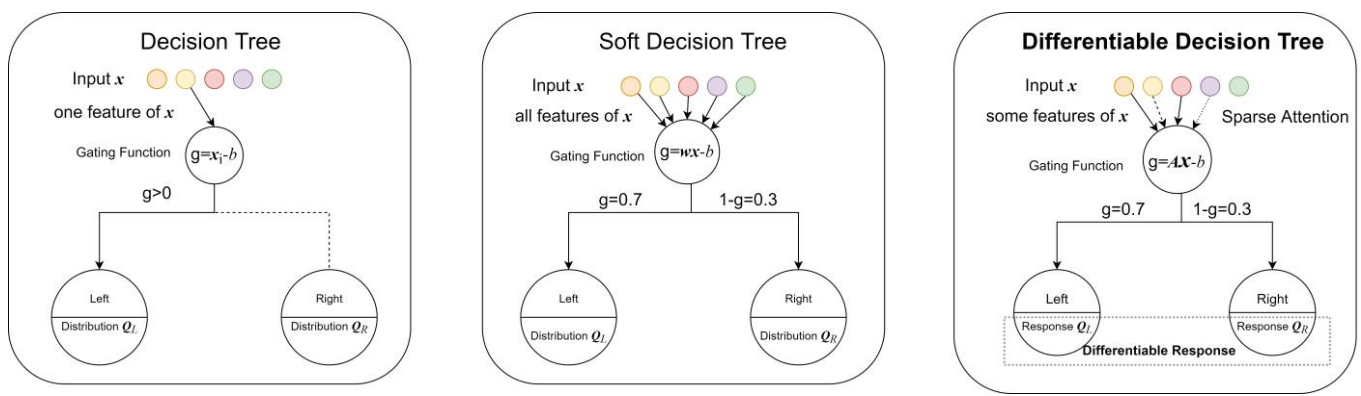

Figure 1 The structure of differentiable decision tree, decision tree and soft decision tree

    As figure 1 shows, differentiable tree has some unique structures compare to other two similar structures: decision tree[4] and soft decision tree[3]. Differentiable tree would learn the most important features to make decision. This would be more reasonable than only on one feature or on all features. Soft decision tree would use all features, which would make the

decision less interpretable and get stuck at local minima. We find the selection of important features would be explained by the attention mechanism. Based on this understanding, we present data-aware initialization technique to improve the sparsity further and get higher accuracy. The general definition and detailed analysis would be given in section 2.

To further study and improve this algorithm, we developed QuantumForest. It's an open-source Python/C++ package and the codes are available at https://github.com/closest-git/QuantumForest. QuantumForest is based on the framework of PyTorch [24]. The main differentiable forest is just a PyTorch module, just like convolution layers of CNN. So it would be easy to combine with any other modules in PyTorch. For example, we would combine the differentiable forest and CNN to study more problems.

## 2. Differentiable Forest with sparse attention

To describe the problem more concisely, we give the following formulas and symbols:

For a dataset with N samples $\mathbf{X} = \{x\}$ and its target $\mathbf{Y} = \{y\}$. Each $x$ has M attributes, $x = [x_1, x_1, \cdots, x_M]^T$. The differentiable forest model would learn an ensemble of differentiable decision trees $\{T^1, T^2, T^3, \ldots, T^K\}$ to minimize the loss between the target $y$ and prediction $\hat{y}$.

$$\hat{y} = \frac{1}{K}\sum_{h=1}^{K} T^h(x) \qquad (1)$$

Figure 1 shows the simplest case of differentiable tree. It has only one gating function $g$ to test the input $x$. There are three nodes. The root node represented the gating function with some parameters. It has two child nodes, which are represented as {left, right} or {↙, ↘}. The root node would redirect input $x$ to both {left, right} with probabilities calculated by the gating function $g$. Formula (2) gives the general definition of gating function, where $A \in R^M$ is a learnable weight parameter for each attribute of $x$, $b$ is a learnable threshold. $\sigma$ would map $Ax - b$ to probability between [0,1], for example, the sigmoid function.

$$g(A, x, b) = \sigma(Ax - b) \qquad (2)$$

So as shown in figure 2, The sample $x$ would be directed to each nodal $j$ with probability $p_j$. And finally, the input $x$ would reach all leaves. For a tree with depth $d$, we represent the path as $\{n_1, n_2, \cdots, n_d\}$, where $n_1$ is the root node and $n_d$ is the leaf node $j$. $p_j$ is just the product of the probabilities of all nodes in this path:

$$p_j = \prod_n g_n \quad \text{where } n \in \{n_1, n_2, \cdots, n_d\} \quad (3)$$

It's the key difference with the classic decision trees, in which the gating function $g$ is just the heave-side function, so each $x$ always get only one state, either left or right. And finally, the input $x$ would only reach one leaf.

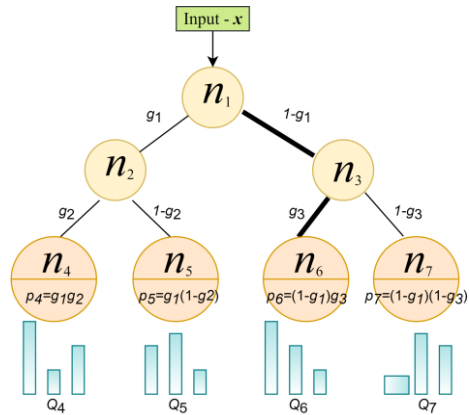

Figure 2 differentiable tree and its probability and response at leaf nodes.
In this sample, The input $x$ would reach $n_3$ with probability 1-$g_1$ and reach $n_6$ with probability (1-$g_1$) $g_3$

Let the response at leaf node *j* represented by learnable parameter $Q_j$. Then the output of the tree is just the probability average of these responses.

$$\hat{y}(x) = \sum_j p_j Q_j(x) \quad \text{where } j \text{ is the leaf} \quad (4)$$

A single tree is a very weak learner, so we should merge many trees to get higher accuracy, just like the random forest or other ensemble learning method. Let $\Theta$ represents all parameters $(A, b, Q)$, then the final loss would be represented by the general form of (5)

$$L(\Theta: x, y) = \frac{1}{K}\sum_{h=1}^{K} L^h(\Theta: x, y) = \frac{1}{K}\sum_{h=1}^{K} L^h(A, b, Q: x, y) \quad (5)$$

In the case of classification problem, the classical function of *L* is cross-entropy. For regression problem, *L* maybe mse, mae, huber loss or others. To minimize the loss (5), we use stochastic gradient descent(SGD) method to train this model.

**Data-aware sparse attention**

As formula (2) shows, $A \in R^M$ set different weight to different attribute of $x$. As the learning goes on, some weights would be bigger and bigger and others would be smaller. So just like the attention mechanisms in deep learning, $A$ is also an attention vector, which would select the most distinguishing attributes and ignore the others. Entmax [9] is a powerful transformer to produce sparse attention. In theory, entmax would assign nearly zero or just zero weight to irrelevant attributes. But its really hard to achieve high sparsity for the huge datasets from practical application. In our experiment, entmax would set zero weights to some attributes. There are still many irrelevant attributes with non zero weights. Based on our knowledge on the application and datasets, there are always some attributes that have higher importance than others. So we would use some fast method to measure the importance of each attribute, which would help gating function to pay more attention on the important attributes. As formula (6) shows:

$$g(A, x, b) \Rightarrow g(\varepsilon(A), x, b) \Rightarrow g(\varepsilon(\ddot{A}), x, b) \quad (6)$$

where ε is entmax function, A initialized from random value and $\ddot{A}$ initialized from the importance of each attribute.

The key is to estimate the importance of each attribute. Modern GBDT libs(LightGBM[6], CatBoost[7] and LiteMORT[8]) provides very fast ways to estimate the importance. Even for dataset with million samps, they would find some estimation in minutes or even seconds. This time is negligible compared with the long time of training process. And in real application, the main difficulty of GBDT method is from the large memory need by these methods, which would always load all datasets to memory and need more temporary memory. LiteMORT would use much less memory than other libs. So we would use LiteMORT to get the importance of each feature. With this data-aware initialization technique, we would get a more sparse attention vector $A$ and get higher accuracy.

**Learning algorithm**

Based on the general loss function defined in (5), we use stochastic gradient descent method [11,25] to reduce the loss. As formula (7) shows, update all parameters $\Theta$ batch by batch:

$$\Theta^{(t+1)} = \Theta^t - \eta \frac{\partial L}{\partial \Theta}(\Theta^t; \mathcal{B}) = \Theta^t - \frac{\eta}{|\mathcal{B}|} \sum_{(x,y) \in \mathcal{B}} \frac{\partial L}{\partial \Theta}(\Theta^t; x, y) \quad (7)$$

where $\mathcal{B}$ is the current batch, $\eta$ is the learning rate, $(x, y)$ is the sample in current batch.

This is similar to the training process of deep learning. Some hyperparameters (batch size, learning rate, weight decay,drop out ratio…) need to be set. All the training skills and experience from deep learning could be used. For example, the batch normalization technique, drop out layer, dense net…We find QHAdam[25] would get a few higher accuracy than Adam[11]. So QHAdam is the default optimization algorithm in QuantumForest.

```
Input
    training, validing and testing dataset
Initialization
    Init the sparse attention A
    Init response Q at each leaf nodes
    Init threshhold values b
repeat
    For each batch B ⊆ X:
        Forward and get loss
```
$$L(\Theta:x,y) = \frac{1}{K}\sum_{h=1}^{K} L^h(A,b,Q:x,y)$$
```
        Backpropagate to get the gradient
            (ΔA, Δb, ΔQ) = ΔL
        Update the parameters
            A ← A − ηΔ
            b ← b − ηΔb
            Q ← Q − ηΔQ
        Evalue loss at validing dataset
until convergence
```

Algorithm 1 Learning differentiable decision trees

## 3. Results and discussion

To verify our model and algorithm, we test its performance on five large datasets. Table 1 lists the detail information of these datasets. We split each dataset into training and testing sets. The training set is used to learn differentialble forest models. The testing set is used to evaluate the performance of the learned models.

Table 1  Five large tabular datasets

|  | Higgs [16] | Click [17] | YearPrediction [18] | Microsoft [19] | Yahoo [20] |
|---|---|---|---|---|---|
| Train | 10.5M | 800K | 463K | 723K | 544K |
| Test | 500K | 200K | 51.6K | 241K | 165K |
| Features | 28 | 11 | 90 | 136 | 699 |
| Problem | Classification | Classification | Regression | Regression | Regression |
| Description | UCI ML Higgs | 2012 KDD Cup | Million Song Dataset | MSLR-WEB 10k | Yahoo LETOR dataset |

### 3.1 Accuracy

We compare the accuracy of QuantumForest with the following libraries:

1) Catboost [7]. A GBDT library which uses oblivious decision trees as weak learners. We use the open-source implementation at https://github.com/catboost/catboost.

2) XGBoost [5]. We use the open-source implementation at https://github.com/dmlc/xgboost

3) NODE [2]. A new neural oblivious decision ensembles for deep learning. We use the open-source implementation at https://github.com/Qwicen/node

4) mGBDT [21]: Multi-layered gradient boosting decision trees by [21]. We use the open-source implementation at https://github.com/Qwicen/node

Catboost and XGBoost are the best GBDT libs, which are the state-of-the-art tools for the tabular datasets. NODE is also based on the differentiable oblivious forest, which is a special version of our model. That is, the nodes in each layer share only one gating function.

Table 2 Accuracy comparison with other models*

|  | Higgs | Click | YearPrediction | Microsoft | Yahoo |
| --- | --- | --- | --- | --- | --- |
| CatBoost [7] | 0.2434 | 0.3438 | 80.68 | 0.5587 | 0.5781 |
| XGBoost [5] | 0.2600 | 0.3461 | 81.11 | 0.5637 | 0.5756 |
| NODE [2] | 0.2412 | 0.3309 | 77.43 | 0.5584 | 0.5666 |
| mGBDT [21] | OOM | OOM | 80.67 | OOM | OOM |
| QuantumForest | **0.237** | **0.3318** | **74.02** | **0.5568** | **0.5656** |

*Some results are copied form the testing results of NODE [2].

Table 2 listed the accuracy of all libraries. All libraries use default parameters. For each dataset, QuantumForest uses 2048 trees, and the batch size is 512. It's clear that mGBDT needs much more memory than other libs. mGBDT always failed because out or memory for most large datasets. Both NODE and QuantumForest have higher accuracy than CatBoost and XGBoost. It is a clear sign that differentiable forest model has more potential than classical GBDT models.

It seems that QuantumForest is better than other libraries on all datasets. We are busy developing some new algorithms and sure its performance would increate a lot. That doesn't mean QuantumForest would be best in all cases. As the famous no free lunch theorem, some lib would perform better in some datasets and maybe poor in other datasets. Anyway, QuantumForest shows the great potential of differentiable forest model. We would report more detailed comparative results in the following papers.

**3.2 Sparse Attention from data-aware initialization**

We use attention heatmap to study the distribution of attention values. Figure 3 is the sample from the YearPrediction problem [18]. The height is the number of attributes.The width is the number of attention vectors. Each pixel represents an attention value. Each pixel represents an attention value. And the colorbar shows the range of all values. In this problem, each $x$ has 90 attributes, so the heighth of each heatmap is always 90. But the size of attentinon vector is very large. So we random pick 120 vectors from some nodes. The width of each heatmap is 120. We compare the heatmap from different initialization technique. The left figure is the heatmap from random initialization. The right figure is the heatmap from data-aware initialization. It's clear that right figure are much sparse than left figure. So each decision is made with fewer and more important attributes. This verified that data-aware initialization technique would make the attention vector more sparse, which would improve the accuracy.

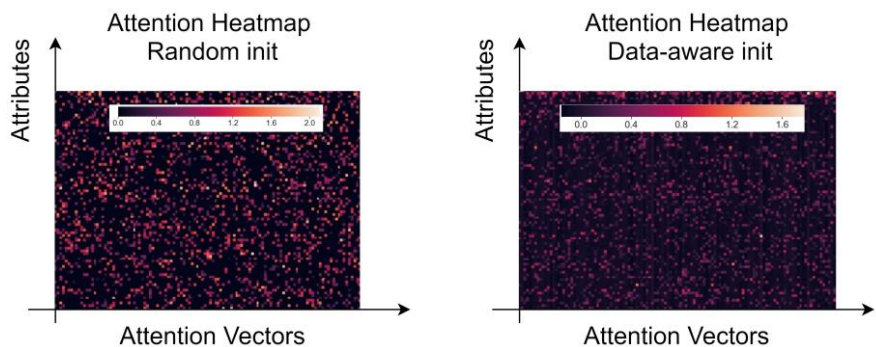

Figure 3 Comparison of attention values from different initialization
The height is the number of attributes.The width is the number of attention vectors. Each pixel represents an attention value

**4 Prospect**

Differentiable forest is a new method and has great potential for more problems. We would try more complex structures. For example, combine the differentiable forest with deep CNN to improve its performance.